# Chapter 1
# Dual-Stage Global and Local Feature Framework for Image Dehazing

Anas M. Ali[0000−1111−2222−3333], Anis Koubaa[1111−2222−3333−4444], and Bilel Benjdira[2222−3333−4444−5555]

**Abstract** Addressing the challenge of removing atmospheric fog or haze from digital images, known as image dehazing, has recently gained significant traction in the computer vision community. Although contemporary dehazing models have demonstrated promising performance, few have thoroughly investigated high-resolution imagery. In such scenarios, practitioners often resort to downsampling the input image or processing it in smaller patches, which leads to a notable performance degradation. This drop is primarily linked to the difficulty of effectively combining global contextual information with localized, fine-grained details as the spatial resolution grows. In this chapter, we propose a novel framework, termed the Streamlined Global and Local Features Combinator (SGLC), to bridge this gap and enable robust dehazing for high-resolution inputs. Our approach is composed of two principal components: the Global Features Generator (GFG) and the Local Features Enhancer (LFE). The GFG produces an initial dehazed output by focusing on broad contextual understanding of the scene. Subsequently, the LFE refines this preliminary output by enhancing localized details and pixel-level features, thereby capturing the interplay between global appearance and local structure. To evaluate the effectiveness of

---

Anas M. Ali

Robotics and Internet-of-Things Laboratory, Prince Sultan University, Riyadh, 12435, Saudi Arabia
e-mail: `aaboessa@psu.edu.sa`

Anas M. Ali

Department of Electronics and Electrical Communications Engineering, Faculty of Electronic Engineering, Menoufia University, Menouf, 32952, Egypt

Anis Koubaa

College of Engineering, Alfaisal University, Riyadh, 11533, Saudi Arabia
e-mail: `akoubaa@alfaisal.edu`

Bilel Benjdira

Robotics and Internet-of-Things Laboratory, Prince Sultan University, Riyadh, 12435, Saudi Arabia
e-mail: `bbenjdira@psu.edu.sa`

Bilel Benjdira

SE & ICT Laboratory, LR18ES44, ENICarthage, University of Carthage, Tunis, 1054, Tunisia

SGLC, we integrated it with the Uformer architecture, a state-of-the-art dehazing model. Experimental results on high-resolution datasets reveal a considerable improvement in peak signal-to-noise ratio (PSNR) when employing SGLC, indicating its potency in addressing haze in large-scale imagery. Moreover, the SGLC design is model-agnostic, allowing any dehazing network to be augmented with the proposed global-and-local feature fusion mechanism. Through this strategy, practitioners can harness both scene-level cues and granular details, significantly improving visual fidelity in high-resolution environments. Overall, this chapter provides an in-depth exposition of the SGLC framework, highlighting its fundamental concepts, implementation details, and empirical validation. Our findings underscore the importance of synergizing global and local features for effective high-resolution image dehazing, paving the way for future research directions and practical applications in domains where clarity and detail are paramount.

## 1.1 Introduction

Image dehazing, which aims to remove the atmospheric fog or haze present in digital imagery, has steadily attracted scholarly attention due to its direct impact on numerous computer vision applications. Tasks such as image classification, object detection, tracking, and semantic segmentation are particularly susceptible to performance degradation when haze is present in the scene. The fundamental challenge lies in handling the complex and non-homogeneous nature of haze particles, which causes uneven scattering and attenuation of light. Consequently, different regions of an image suffer from varying degrees of degradation and blur, rendering traditional computer vision pipelines less effective.

Early research endeavors in this field often concentrated on simpler scenarios of homogeneous haze (1; 33; 5; 19). However, realistic environments typically contain spatially varying haze distributions, thereby motivating the need for more sophisticated models. The widely accepted physical model for describing the propagation of light in hazy conditions can be formulated by:

$$I_h(x) = t(x)\, I_o(x) + \bigl(1 - t(x)\bigr) A(x), \tag{1.1}$$

where $I_h$ denotes the observed hazy image, $I_o$ is the latent (clean) scene, $t(x)$ is the medium transmission map capturing the haze-induced degradation, and $A(x)$ signifies the global atmospheric light (9; 24; 40). The transmission map $t(x)$ is generally expressed as:

$$t(x) = e^{-\beta\, d(x)}, \tag{1.2}$$

where $\beta$ is an atmospheric scattering parameter and $d(x)$ represents the scene depth as a function of pixel coordinates $x$.

Before the advent of deep learning, many studies relied on prior-based techniques (26; 18; 14) to approximate $\beta$, $A(x)$, and $t(x)$. Methods such as the dark channel prior and non-local prior were employed to derive estimations of the atmospheric scattering effects. However, these hand-crafted priors often fell short in handling the intricate dependencies between haze, depth, and environmental factors. The introduction of deep learning methods (35) brought about substantial improvements in haze removal performance. Subsequent models based on convolutional neural networks (CNNs) or transformer-based architectures demonstrated remarkable accuracy gains (29; 34; 6; 43; 36; 23; 38).

Despite these achievements, a significant limitation persists: state-of-the-art dehazing models are typically designed for small to medium input resolutions, making them computationally expensive or less effective when dealing with high-resolution images (42; 17). Incorporating advanced network components (e.g., vision transformers) further escalates computational demands (44; 37), and attempts to balance performance and model efficiency pose ongoing challenges (2; 10; 31). One of the central difficulties lies in effectively fusing global scene context with localized details in large-scale images. While low-resolution inputs can be processed in a single pass, high-resolution images often require downsampling or patch-wise processing. Such approaches risk losing crucial global features or fine-grained local details, thereby impairing the model's overall performance.

In this book chapter, we introduce the *Streamlined Global and Local Features Combinator* (SGLC) to enable more robust dehazing for high-resolution imagery. SGLC facilitates an end-to-end process that preserves both the global scene context and the local fine structure. Specifically, the method comprises two main components:

- **Global Features Generator (GFG):** Responsible for capturing broad, scene-level characteristics by operating on strategically subdivided patches from the high-resolution input.
- **Local Features Enhancer (LFE):** Dedicated to refining and augmenting localized details, thus ensuring the recovery of subtle edges, textures, and other critical high-frequency components.

Additionally, we introduce a Grid Patching process tailored for high-resolution images to capture their spatial variability more effectively. We also propose a customized loss function aimed at reinforcing the learning of high-frequency details, vital for visually pleasing and accurate dehazed outcomes. Finally, we confirm through comparative experiments that applying GFG prior to LFE aligns with the intuitive approach of first generating holistic context before focusing on fine detail enhancement.

Overall, this chapter addresses the persistent challenge of balancing global and local feature extraction in the dehazing of large-scale images. By systematically combining robust global scene understanding with fine-grained local refinements, the SGLC framework offers a scalable and computationally feasible pathway to high-resolution image dehazing, thereby broadening the applicability and impact of modern dehazing techniques.

## 1.2 Related Work

### 1.2.1 Background on Image Dehazing

Image dehazing, a critical pre-processing step for many computer vision tasks, has been extensively studied in both academic and industrial contexts. Early approaches generally used physical priors to estimate haze-relevant parameters. For instance, the dark channel prior (16) and the non-local prior (3) are classic techniques that derive statistics from hazy scenes to approximate the transmission map and atmospheric light. Other notable methods focused on modeling the interactions of light with haze particles via sophisticated atmospheric scattering equations (13; 15). Although these hand-crafted strategies proved effective under certain assumptions, they often struggled to accommodate complex real-world conditions involving non-homogeneous and dense haze distributions.

With the advent of deep learning, data-driven dehazing models have rapidly gained traction and demonstrated superior performance. Early convolutional neural network (CNN) approaches (7; 28) replaced explicit priors with learned feature representations, allowing models to adapt to more diverse haze conditions. Further developments incorporated advanced architectures such as generative adversarial networks (GANs) (30; 20) and multi-scale designs (25; 27; 39) to handle varying levels of haze density and more complex scene structures. Moreover, several frameworks have pursued end-to-end training by integrating multiple loss terms (e.g., reconstruction loss, perceptual loss) to enhance image quality in both global appearance and fine details (21; 22; 12; 8).

### 1.2.2 High-Resolution Image Dehazing

Although substantial progress has been made in single-image and multi-image dehazing for standard resolutions, the field has increasingly turned its attention to the challenges posed by large-scale, high-resolution imagery. When attempting to extend existing methods to high-resolution inputs, computational costs can become prohibitive, often necessitating either image downsampling or a patch-wise processing strategy. These approaches, however, risk discarding valuable global context and degrading local details.

Among the first attempts to specifically address the high-resolution dehazing problem, Sim et al. (30) proposed the Dehazing Generative Adversarial Network (DHGAN). Their method trains a generator on smaller hazy patches derived from downsampled input images, thereby reducing computational overhead and capturing essential global cues. To further strengthen the training process, they modified the cross-entropy loss to accommodate multiple outputs. Similarly, Ki et al. (20) introduced BEGAN (Boundary Equilibrium Generative Adversarial Network), which uses an enlarged receptive field and focuses on training the discriminator directly on high-resolution images, while conditioning the generation process on downscaled hazy counterparts.

In a related effort, Bianco et al. (4) presented HR-Dehazer (High-Resolution Dehazer), an encoder-decoder-based model combined with a specialized loss function that prioritizes semantic integrity and structural consistency. The network design is scale-invariant, enabling operation on large inputs without a considerable drop in accuracy. More recently, Zheng et al. (41) offered 4k-Dehazer, a model that employs three subnetworks working in tandem on a bilateral space. Each branch shares features bidirectionally, producing a robust global representation suitable for 4K images. To evaluate their system, the authors introduced a large 4K dataset, showing competitive results against contemporary methods.

Another strategy for high-resolution dehazing is illustrated by Chen et al. (8), who proposed H2RL-Net, a two-branch framework that separately addresses high-frequency and semantic information. By leveraging multi-resolution CNN streams along with a parallel cross-scale fusion (PCF) module, H2RL-Net dynamically integrates features across different scales, while the channel feature refinement (CFR) block recalibrates these features at the channel level. This division of labor aids in capturing both small-scale details and global context.

### 1.2.3 Limitations of Existing Approaches and Our Contribution

Although these high-resolution dehazing methods have reported notable gains, they generally rely on designing new architectures specifically tailored for large images. This tends to limit practitioners who may prefer to use well-established models designed for mid-resolution dehazing tasks. Moreover, most of the aforementioned techniques integrate the extraction of global and local features in a parallel manner, whether through bilateral latent spaces or concurrent multi-scale networks. While such parallel designs have proven effective, they can be computationally heavy and less modular for incremental improvements.

In this chapter, we address the gap between existing high-resolution dehazing methods and the extensive body of CNN- or transformer-based dehazing models. We propose a sequential, streamlined approach called *Streamlined Global and Local Features Combinator (SGLC)* to preserve both global scene semantics and fine-grained local details without demanding substantial architecture overhauls. Specifically, Our framework centers on two key modules. The first, termed the **Global Features Generator (GFG)**, processes large images via a Grid Patching mechanism to capture essential global statistics and contextual cues. The second, called the **Local Features Enhancer (LFE)**, refines the resulting representations by restoring high-frequency details and enhancing local structure, thereby boosting overall perceptual quality.

The sequential nature of SGLC simplifies the evaluation and potential refinement of each stage independently, offering a flexible and extensible architecture. By decoupling global feature generation from local feature enhancement, our method avoids common pitfalls associated with patch-wise downsampling and large memory footprints. Consequently, SGLC serves as a generic and versatile framework that can be integrated into existing state-of-the-art dehazing models, extending their applicability to high-resolution settings with minimal modifications and enhanced performance.

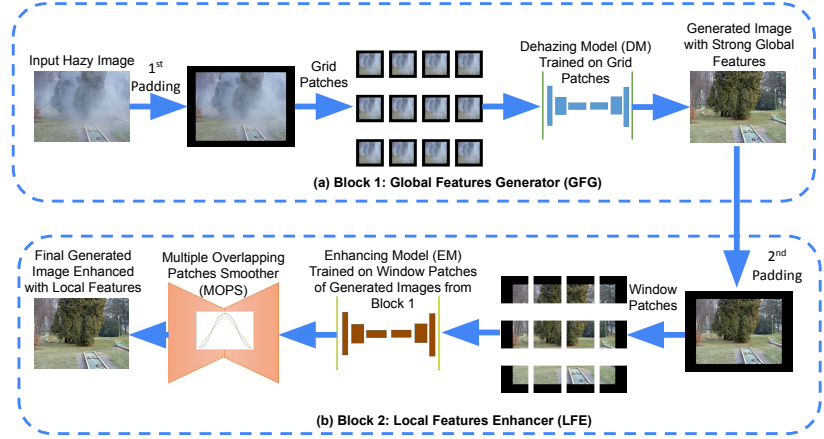

**Fig. 1.1** SGLC diagram.

## 1.3 Proposed Methodology

In this section, we present the core components of our *Streamlined Global and Local Features Combinator (SGLC)* framework, whose primary objective is to enable high-resolution image dehazing while preserving both global scene context and fine local details. Figure 1.1 provides an overview of the proposed methodology, illustrating the two main blocks that compose the SGLC pipeline: the *Global Features Generator (GFG)* and the *Local Features Enhancer (LFE)*. In what follows, we describe each module and its associated algorithms in detail.

### 1.3.1 Global Features Generator (GFG)

The first stage of the SGLC framework, termed the Global Features Generator (GFG), aims to capture and reconstruct the essential global appearance of the hazy image. High-resolution images impose significant memory and computational burdens, making it impractical to process them holistically in a single forward pass. To address this issue, we introduce a *Grid Patching* mechanism that systematically samples the full-resolution image with minimal overlap, thereby maintaining a coarse-level representation of the entire scene.

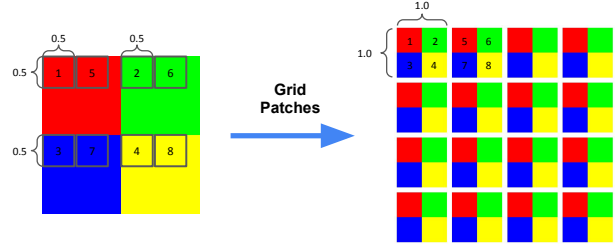

**Fig. 1.2** Grid patches diagram.

#### 1.3.1.1 Grid Patching for Global Features Learning

Consider an input image $I$ of height $H$ and width $W$. Our approach begins by introducing padding to ensure that $H$ and $W$ become divisible by a patch size $G \times G$. We denote the padded image by $I'$ with height $H'$ and width $W'$. If $H$ (respectively $W$) is already divisible by $G$, then $H' = H$ (respectively $W' = W$); otherwise, $H'$ and $W'$ are computed as follows:

$$H' = \left(\lfloor H/G \rfloor + 1\right) \times G, \quad W' = \left(\lfloor W/G \rfloor + 1\right) \times G. \tag{1.3}$$

We define the number of vertical divisions as $n_h = H'/G$ and the number of horizontal divisions as $n_w = W'/G$. The total number of patches is then given by

$$N = n_h \times n_w. \tag{1.4}$$

Algorithm 1 outlines the Grid Patching procedure. Each generated patch, denoted $P_k$ for $k = 0, \ldots, N-1$, is an array of size $G \times G \times 3$ (assuming RGB color channels). The central idea is to sample pixels from the padded image $I'$ at intervals of $n_h$ along the vertical axis and $n_w$ along the horizontal axis. By doing so, the patch content collectively spans the entire image, capturing a broad overview of the scene and facilitating the learning of global features, as shown in Figure 1.2.

**Algorithm 1** Grid Patching

**Require:** $I'$: The padded image, $n_h$: The height number of divisions, $n_w$: The width number of divisions, $N$: The number of generated patches
**Ensure:** A set of grid patches $\{P_k\}_{k=0}^{N-1}$
1: **for** $k = 0$ to $N - 1$ **do**
2:    $P_k \leftarrow \text{zeros}(G, G, 3)$　　　　　　　　　　　　　　▷ Initialization to zeros
3:    **for** $i = 0$ to $G - 1$ **do**
4:       **for** $j = 0$ to $G - 1$ **do**
5:          $P_k[i, j] \leftarrow I'[i + n_w, j + n_h]$
6:       **end for**
7:    **end for**
8: **end for**

#### 1.3.1.2 Dehazing Model (DM)

Once the grid patches are obtained, they are individually fed to a dehazing model (DM), which is designed to learn scene-level information from these patches. In this work, we adopt Uformer (32) as our baseline dehazing architecture, although SGLC is by no means limited to this particular choice. Uformer is built on a hierarchical U-shaped network with skip connections reminiscent of the classic U-Net structure. By incorporating LeWin Transformer blocks, Uformer captures both short-range and long-range dependencies using a locally enhanced window-based attention mechanism.

More concretely, let $i_h \in \mathbb{R}^{c \times h \times w}$ be a hazy patch. Uformer begins with a $3 \times 3$ convolution and LeakyReLU activation to derive initial feature maps, followed by several encoder stages each containing (1) a stack of LeWin Transformer blocks and (2) a down-sampling layer. The down-sampling layer halves the spatial dimensions while expanding the channel dimension. A decoder mirrors the encoder, employing up-sampling and additional LeWin Transformer blocks, as well as skip connections that fuse low- and high-level features across scales. The local attention windows, of size $m \times m$, break the potentially large feature maps into smaller non-overlapping regions, thereby reducing the self-attention computational overhead. Although originally designed to combine both global and local features within moderate-resolution imagery, Uformer (and similar architectures) can lose its capacity to holistically encode global context when scaled to very large image sizes. SGLC resolves this issue by systematically providing global coverage through Grid Patching.

#### 1.3.1.3 Self-Supervised Learning

Inspired by works such as (1; 33; 5; 19), we employ a preliminary self-supervised training step for both the Dehazing Model (DM) and the subsequent Enhancer Model (EM). Specifically, we generate an auxiliary dataset in which small squares of each clean image are randomly in-painted with white patches, simulating localized hazy regions. The model is trained to reconstruct the original, uncorrupted image from

1 Dual-Stage Global and Local Feature Framework for Image Dehazing9these artificially degraded samples. This procedure compels the network to capture the semantic context necessary for accurately filling in missing or obscured details, thereby enhancing its ability to handle realistic haze distributions.#### 1.3.1.4 Customized Loss Function

To train both the DM and the EM, we utilize a customized loss function that balances spatial fidelity, structural consistency, and robustness. In particular, we define

$$\mathcal{L} = \sqrt{\|I - \hat{I}\|^2 + \|\Pi(I) - \Pi(\hat{I})\|^2 + \varepsilon^2}, \quad (1.5)$$

where $I$ denotes the ground-truth image, $\hat{I}$ is the predicted output, and $\Pi(\cdot)$ represents the Laplacian Pyramid operator (9). The term $\|I - \hat{I}\|^2$ enforces pixel-level consistency between the prediction and the ground truth. Meanwhile, $\|\Pi(I) - \Pi(\hat{I})\|^2$ emphasizes the alignment of high-frequency details by comparing the Laplacian Pyramid representations of the images (24). The small constant $\varepsilon$ (set to $10^{-3}$ in our experiments) is included following the Charbonnier penalty function (40), making the training process more stable and less prone to vanishing gradients in the regime where $\|I - \hat{I}\|$ is small. Similar strategies have demonstrated enhanced robustness in various image restoration tasks (26).

#### 1.3.1.5 Reverse Grid Patches Reconstruction

After the DM has processed each grid patch $\hat{P}_l$ (for $l \in \{0, \ldots, N-1\}$), we merge the predictions back into a single full-size image $\hat{I}_F$. This step, referred to as *Reverse Grid Patches Reconstruction*, is summarized in Algorithm 2. Initially, we initialize an empty array of the target size $G \times n_w \times G \times n_h$. We then assign each patch $\hat{P}_l$ to its corresponding spatial position in $\hat{I}_F$ based on horizontal and vertical offsets computed from $l$. Finally, the image is cropped to its original dimensions (removing any padding). The resulting image, $\hat{I}_F$, forms the first-stage dehazed output produced by GFG.

### 1.3.2 Local Features Enhancer (LFE)

While the Global Features Generator (GFG) described in the previous subsection provides a coherent large-scale representation of the scene, additional refinement is often required to restore subtle textures, edges, and other small-scale details. In the proposed *Streamlined Global and Local Features Combinator* (SGLC) framework, this role is fulfilled by the *Local Features Enhancer (LFE)*, which focuses on rectifying any deficiencies left unresolved by the GFG output, as shown in Figure 1.3.

**Algorithm 2** Reverse Grid Patches Reconstruction

**Require:** Predicted patches $\{\hat{P}_l\}_{l=0}^{N-1}$; grid size $G$; number of vertical divisions $n_h$; number of horizontal divisions $n_w$; total number of patches $N$.
**Ensure:** Reconstructed image $\hat{I}_F$.
1: $\hat{I}_F \leftarrow \text{zeros}(G \cdot n_w, G \cdot n_h, 3)$
2: **for** $l \in \{0, \ldots, N-1\}$ **do**
3:     $i_0 \leftarrow l \bmod n_w$   // horizontal offset
4:     $j_0 \leftarrow l \div n_w$   // vertical offset
5:     **for** $i_P \in \{0, \ldots, G-1\}$ **do**
6:         **for** $j_P \in \{0, \ldots, G-1\}$ **do**
7:             $x \leftarrow i_0 \times G + i_P$
8:             $y \leftarrow j_0 \times G + j_P$
9:             $\hat{I}_F[x, y] \leftarrow \hat{P}_l[i_P, j_P]$
10:         **end for**
11:     **end for**
12: **end for**

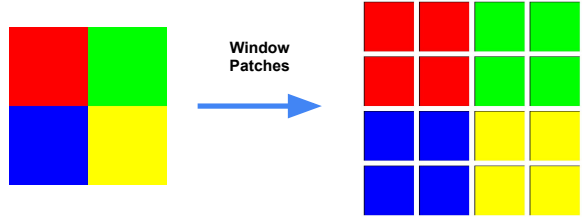

**Fig. 1.3** window patches diagram.

#### 1.3.2.1 Enhancer Model (EM)

Compared to the Dehazing Model (DM) used in the GFG stage, the Enhancer Model (EM) has two key differences. First, it operates on *window patches* (rather than grid patches) to capture spatially continuous regions. Second, it is trained on a new dataset composed of the GFG-generated dehazed images alongside their corresponding clean counterparts. Concretely, for each training sample, the GFG network is first used to generate an initial dehazed version of the hazy input. Both the resulting dehazed image and the corresponding ground-truth clean image are then padded to ensure divisibility by the chosen window size. Next, both padded images are subdivided into patches. These window patches form a new training set for the EM, which learns to focus on fine-grained improvements not fully resolved by the coarse-grained global predictions.

As in the GFG block, we employ the Uformer network (32) for the EM architecture, maintaining the same self-supervised initialization and the same customized loss function introduced earlier (Section 1.3). By following this multi-step training approach, the EM can better capture local feature nuances and ameliorate minor errors introduced by the GFG predictions. This strategy effectively transforms the dehazing process into two complementary stages, where coarse global restoration is tackled first, and fine local refinement is addressed second.

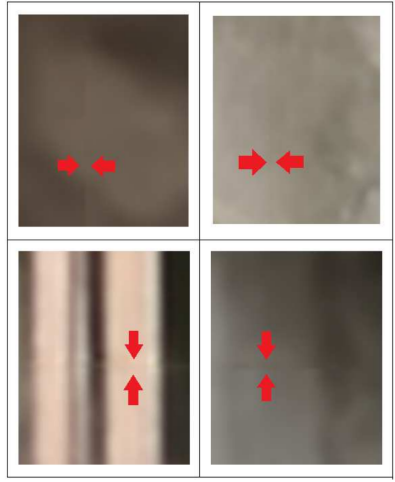

**Fig. 1.4** Vignetting artifacts encountered during the application of Expectation Maximization (EM) in conjunction with the Reverse Window Patches Reconstruction algorithm.

#### 1.3.2.2 Multiple Overlapping Patches Smoother (MOPS)

A naive approach to reconstruct the final image from window patches—using a simple one-to-one arrangement—can introduce visible seams or discontinuities, commonly referred to as *vignetting artifacts* (see Figure 1.4). These artifacts arise because convolutional neural networks (CNNs) often exhibit limited translational equivariance, a limitation exacerbated by zero-padding and strided operations (1). Although such artifacts may be subtle, they can negatively impact both perceptual quality and quantitative dehazing metrics.

To mitigate these discontinuities, we adapt a *blending algorithm* (33) based on overlapping patch predictions and spatial window functions (5). In the context of this study, we refer to this strategy as the *Multiple Overlapping Patches Smoother (MOPS)*. Specifically, the MOPS procedure aggregates multiple overlapping patches, each predicted by the EM under slight variations in the input (e.g., through rotations or reflections). By summing these predictions with a second-order spline window function, MOPS achieves a more seamless blending of local patch reconstructions and reduces boundary inconsistencies. Originally proposed for improving semantic segmentation results (33; 5), we adapt this technique here to enhance high-resolution dehazing outputs.

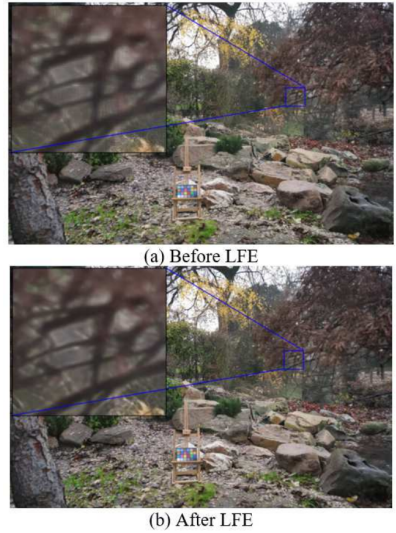

**Fig. 1.5** The predicted image before and after applying the Local Feature Enhancement (LFE) block.

While MOPS significantly improves local consistency, its computational overhead is correspondingly increased. Therefore, Section 1.4 includes an ablation study evaluating the trade-off between the performance gains offered by MOPS and the additional computational cost incurred.

### 1.3.2.3 Visual Illustration and Final Output

After applying MOPS, the EM produces the final, refined dehazed image, which is expected to feature noticeably improved local detail and fewer perceptual artifacts compared to the GFG-only prediction. Figure 1.5 provides a qualitative comparison of a scene before and after local feature enhancement, underscoring the increased sharpness and clarity. Furthermore, the quantitative advantages of the LFE stage are evidenced in the metric evaluations described later, indicating that the two-stage SGLC pipeline delivers superior dehazing performance relative to single-stage approaches.

## 1.4 Experimental Evaluation

In this section, we examine the empirical effectiveness of the proposed *Streamlined Global and Local Features Combinator (SGLC)* on high-resolution dehazing tasks. We contrast its performance not only against the original Uformer (32) applied to downscaled images, but also with two additional methods representative of recent state-of-the-art research: *DW-GAN* (24), the winning approach in the 2021 NTIRE Non-Homogeneous Dehazing Challenge (9), and *4K-Dehazer* (42), an algorithm specifically tailored for large images. Notably, many existing methods are forced to perform an aggressive resize owing to GPU memory constraints, thereby sacrificing local detail. By contrast, our SGLC framework is designed to handle large image dimensions more efficiently, preserving both global and local information.

### 1.4.1 Experimental Configurations

#### 1.4.1.1 Dataset Description

We evaluate our approach on the *HD-NH-HAZE* dataset, introduced as part of the *High-Resolution Non-Homogeneous Dehazing Challenge* at NTIRE 2023. This dataset consists of 50 images in extremely large dimensions (either $4000 \times 6000$ or $6000 \times 4000$), of which 40 are designated for training, 5 for validation, and 5 for testing. As the corresponding clean (ground-truth) images for the validation and test splits are withheld by the challenge organizers, we confined ourselves to the 40 images that include both hazy and clean pairs. From these 40 images, we separated out 36 for training and 4 for testing in our local experiments. No supplementary external data were utilized.

#### 1.4.1.2 Implementation Details

All SGLC experiments were performed on a Lambda AI server equipped with 8 NVIDIA QUADRO 8000 GPUs, each providing 48 GB of GDDR6 memory. The server includes 512 GB of RAM and two Intel Xeon Silver 4216 CPUs (16 cores each). Additionally, we carried out some experiments on a single-workstation setup featuring a single NVIDIA QUADRO 8000 GPU. Our pipeline is implemented in `PyTorch`, with typical hyperparameter settings following established dehazing benchmarks (e.g., batch size between 2 and 8, depending on network depth, and initial learning rates in the range $1 \times 10^{-4}$ to $2 \times 10^{-4}$).

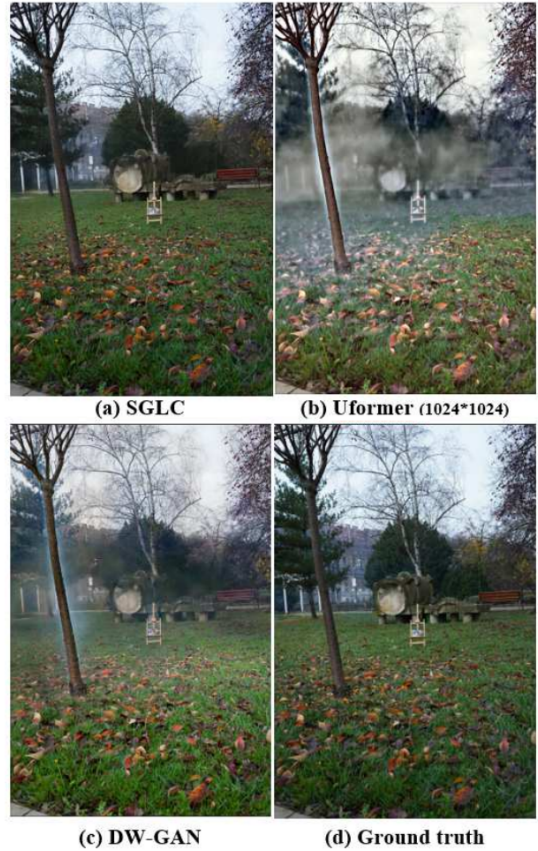

**Fig. 1.6** Dehazing performance of SGLC compared to Uformer (1024*1024), DW-GAN, and the Ground Truth.

### 1.4.2 Results and Analysis

We report quantitative results on the local test subset of 4 images using several standard metrics: Peak Signal-to-Noise Ratio (PSNR) and Structural Similarity Index Measure (SSIM). In addition, we measure the average inference time (in seconds per image) to capture computational overhead. All experiments employ a patch size of $1024 \times 1024$ for both *Grid Patching* in the Global Features Generator (GFG) and *Window Patching* in the Local Features Enhancer (LFE).

Table 1.1 compares our proposed SGLC approach (with different configurations) to Uformer (32), DW-GAN (24; 9), and 4K-Dehazer (42). Uformer was tested on images downscaled to $1024 \times 1024$, as larger dimensions were infeasible due to memory constraints; similarly, 4K-Dehazer was restricted to a maximum resolution of $4000 \times 4000$.

**Table 1.1** Comparison of multiple dehazing methods on HD-NH-HAZE (our local test set). The inference time is computed per image of size $4000 \times 6000$.

| Model Description | PSNR (dB) | SSIM | Inference Time (sec/image) |
| --- | --- | --- | --- |
| Uformer on resized $1024 \times 1024$ | 14.90 | 0.6403 | 14.9 |
| 4K-Dehazer | 17.46 | 0.7230 | 9.0 |
| DW-GAN | 17.48 | 0.7367 | 45.7 |
| **SGLC-GFG only** | 24.49 | 0.8176 | 46.0 |
| **SGLC w/o MOPS** | 25.38 | 0.8511 | 86.0 |
| **SGLC (GFG + LFE + MOPS)** | **25.43** | **0.8524** | 553.5 |
| Inv-SGLC LFE only w/o MOPS | 23.07 | 0.8307 | 40.0 |
| Inv-SGLC LFE only + MOPS | 23.25 | 0.8335 | 556.1 |
| Inv-SGLC (LFE + GFG + MOPS) | 24.39 | 0.8392 | 605.7 |

The resizing approach severely hampers Uformer's ability to capture fine haze structures, leading to relatively poor performance (PSNR 14.90 dB, SSIM 0.6403). Although 4K-Dehazer and DW-GAN improve over this baseline, they do not significantly close the gap. By contrast, our proposed SGLC achieves markedly higher PSNR and SSIM, especially when both the Global Features Generator (GFG) and Local Features Enhancer (LFE) are deployed. We observe that the GFG alone already delivers a substantial boost over the other baselines (PSNR 24.49 dB, SSIM 0.8176). Further local refinement via the LFE block yields an additional improvement of about 0.89 dB in PSNR and 0.0335 in SSIM, The visual comparison in Figure 1.6.

We also investigate the use of MOPS (Multiple Overlapping Patches Smoother), a blending algorithm to reduce patch-based vignetting artifacts. Although its contribution to the final PSNR and SSIM is modest (approximately 0.05 dB in PSNR and 0.0013 in SSIM), it can be critical for scenarios demanding the highest perceptual quality. The ablation study indicates that MOPS accounts for a significant fraction of the total computational cost (about 467.5 seconds out of the total 553.5 seconds per image), suggesting a trade-off between inference time and optimal visual fidelity.

Although the complete SGLC pipeline (including MOPS) requires more than 500 seconds to process a single $4000 \times 6000$ image, this overhead is still feasible for offline or batch-processing settings. If real-time or near-real-time performance is paramount, disabling MOPS reduces total inference to around 86 seconds per image with only a slight dip in performance (SGLC w/o MOPS achieves 25.38 dB PSNR versus 25.43 dB with MOPS).

We further investigate the impact of reversing the order of global and local feature modules. In the inverse setting (Inv-SGLC), the hazy images first pass through a local-focused module (LFE), followed by a global approach (GFG). As indicated in Table 1.1, every configuration of Inv-SGLC underperforms SGLC, highlighting the importance of first capturing global structure before refining local details.

**Table 1.2** Rankings in the NTIRE 2023 Non-Homogeneous Dehazing Challenge (11). "No extra data" refers to solutions trained solely on the official challenge data.

| Model | PSNR | SSIM | LPIPS | MOS |
|---|---|---|---|---|
| **SGLC Performance** | 22.27 | 0.70 | 0.439 | 7.40 |
| **Best "No Extra Data" Solution** | 22.27 | 0.70 | 0.384 | 7.65 |
| SGLC Rank (No Extra Data) | 1/12 | 1/12 | 7/12 | 2/12 |
| **Best Overall Performance** | 22.96 | 0.71 | 0.345 | 8.70 |
| SGLC Rank (Overall) | 3/17 | 4/17 | 10/17 | 5/17 |

**Table 1.3** NTIRE 2023 Non-Homogeneous Dehazing Challenge (11) final leaderboard (with the *Username* column removed). Numbers in parentheses indicate each method's sub-ranking for that specific metric. **SGLC**, which corresponds to our SGLC framework, ranked 5$^{th}$ overall and 2$^{nd}$ among solutions that did not rely on external training data.

| Rank | Models | PSNR↑ | SSIM↑ | LPIPS↓ | Params (M) | Device | Extra Data |
|---|---|---|---|---|---|---|---|
| 1 | DWT-FFC GAN | 22.87 (2) | 0.71 (2) | 0.346 (2) | 373 | RTX2080 Ti | Yes |
| 2 | ITB Dehaze | 22.96 (1) | 0.71 (1) | 0.345 (1) | 110 | 2×TitanXP | Yes |
| 3 | [Mask] | 22.90 (4) | 0.68 (9) | 0.501 (13) | 9.31 | A100 | No |
| 4 | NUSRIQC DEHAZING | 21.97 (8) | 0.68 (9) | 0.380 (3) | 25.58 | 4×RTX3090 | Yes |
| **5** | **SGLC (our)** | **22.49 (6)** | **0.70 (3)** | **0.439 (10)** | **5.0** | **RTX8000** | **No** |
| 6 | NTU607-dehaze | 21.86 (10) | 0.73 (1) | 0.442 (11) | 101.5 | V100 | Yes |
| 7 | MIPCer | 21.75 (12) | 0.69 (6) | 0.464 (8) | 2.39 | A100 | No |
| 8 | iPAL-LightDehaze | 22.09 (9) | 0.67 (13) | 0.556 (16) | n/a | TitanXP | No |
| 9 | Xsourse | 22.09 (13) | 0.65 (12) | 0.556 (16) | n/a | A4000 | No |
| 10 | Xiaofeng Cong | 21.89 (14) | 0.64 (14) | 0.470 (9) | 24.7 | A100 | No |
| 11 | MengFeiHome | 20.96 (15) | 0.62 (15) | 0.515 (14) | n/a | RTX3060 | No |
| 12 | CANT HAZE | 20.95 (16) | 0.69 (6) | 0.415 (6) | 80 | T4/K80 | No |
| 13 | IR-SDE | 19.64 (17) | 0.61 (16) | 0.406 (7) | 78 | A100 | No |

### 1.4.3 Challenge Leaderboard

To verify the consistency of our local results, we also report the final standings from the *NTIRE 2023 Non-Homogeneous Dehazing Challenge* (11), shown in Table 1.2. SGLC placed 5$^{th}$ among 13 submitted solutions when considering multiple metrics (PSNR, SSIM, LPIPS, and MOS). When restricting attention to methods that did not employ extra external training data, SGLC attained the top PSNR and SSIM scores and the second-highest MOS.

Notably, SGLC tied for the best PSNR among entries that did not leverage additional training images, while also sharing the top SSIM score. Among all solutions, SGLC placed third in PSNR (0.69 dB behind first place) and fourth in SSIM (0.01 difference from the leader). The main shortfall lies in the LPIPS metric, hinting at potential avenues for future refinement (e.g., incorporating perceptual loss terms). Overall, these challenge results confirm the effectiveness and adaptability of SGLC in addressing high-resolution, non-homogeneous dehazing with minimal reliance on extra data.

As shown in Table 1.3, **SGLC** (our proposed SGLC framework) attained an overall rank of 5 across multiple metrics, including PSNR, SSIM, LPIPS, and MOS. Moreover, when considering only those solutions that did not rely on any external dataset for training, SGLC placed second in the competition. This performance underscores the ability of the SGLC methodology to accurately remove haze in large-scale, high-resolution images without necessitating additional data sources. Our experiments demonstrate that SGLC substantially improves dehazing performance for large-scale images compared to resizing-based baselines and specialized high-resolution methods. Despite the higher computational cost incurred by MOPS, the overall framework exhibits notable flexibility: one can omit MOPS if inference time is the priority, retaining most of the accuracy gains. Finally, the challenge leaderboard underscores SGLC's competitiveness, particularly in data-constrained settings where access to external training resources is limited.

## 1.5 Conclusion

In this chapter, we introduced the *Streamlined Global and Local Features Combinator* (SGLC) framework, designed to tackle the longstanding challenge of high-resolution image dehazing. By decomposing the dehazing process into two specialized stages—the *Global Features Generator* (GFG) and the *Local Features Enhancer* (LFE)—SGLC effectively addresses both large-scale atmospheric artifacts and fine-grained textual details. The GFG stage capitalizes on a *Grid Patching* strategy, enabling comprehensive scene-level restoration without exceeding computational resources, while the LFE stage adopts *Window Patching* and an overlapping-patch smoothing algorithm to refine localized structures and mitigate vignetting artifacts. Extensive experiments on high-resolution datasets indicate that SGLC consistently achieves superior quantitative and qualitative results compared to conventional single-stage approaches and alternative high-resolution dehazing methods. Our ablation studies reveal that (1) the GFG and LFE modules each independently contribute to performance gains, and (2) the sequential order of applying global and local refinement proves crucial. Additionally, although the Multiple Overlapping Patches Smoother (MOPS) algorithm incurs a computational cost, it can further enhance perceptual fidelity when ultimate image quality is required. Overall, this work highlights the importance of carefully orchestrating global and local feature extraction to handle complex, large-scale haze conditions. The modular nature of SGLC also encourages future extensions, such as integrating more efficient patch division schemes, incorporating advanced Transformer-based blocks, or exploring additional self-supervised pretraining paradigms. By providing a robust and flexible two-stage solution, SGLC serves as a promising framework that pushes the boundaries of high-resolution image dehazing in both research and real-world deployments.

**Acknowledgements** The authors thank Prince Sultan University for the support and funding in conducting this research.

**Competing Interests** The authors declare that they have no competing interests related to this work.

**Ethics Approval** This study does not involve any primary research with human or animal subjects. Therefore, ethical approval was not required.

## Author Biographies

**Anas M. Ali** is a Research Assistant within the Robotics and Internet of Things Laboratory at Prince Sultan University (PSU). He received his Master's Degree in Electrical Engineering from the Faculty of Electronic Engineering, Menoufia University, Egypt, in 2021. His primary research interests include computer vision, healthcare image analysis, deep learning, image restoration, medical image processing, and remote sensing.

**Anis Koubaa** is the Leader and Founder of the Robotics and Internet-of-Things Laboratory at Prince Sultan University. He is also the Executive Director of the Innovation Center at Prince Sultan University. He is a full professor of computer science and has been working on several research and development projects on robotics, unmanned systems, deep learning, and the Internet of Things. He is a Senior Fellow of the Higher Education Academy, U.K. He has presented numerous training programs on cutting-edge technologies. His research interests include developing artificial intelligence products and automated solutions for logistics, using drones and robots for delivery systems.

**Bilel Benjdira** is an Assistant Professor at Prince Sultan University (PSU) and a researcher within the Robotics and Internet of Things Laboratory at PSU. He received his Ph.D. degree in Electrical Engineering from ENICarthage, University of Carthage, Tunisia, in 2021. His primary research interests include computer vision, healthcare image analysis, deep learning, and remote sensing.